\documentclass{article}
\usepackage{spconf,amsmath,graphicx}
\usepackage{url}
\usepackage{subfigure}
\usepackage{algorithm}
\usepackage{algorithmic}
\usepackage[algo2e,ruled,linesnumbered]{algorithm2e}
\usepackage{etoolbox}
\usepackage{multirow}
\usepackage{threeparttable}
\usepackage{balance}
\usepackage{booktabs} 

\usepackage[hidelinks]{hyperref}

\usepackage{color}
\usepackage{xcolor,soul}

\newcommand{\tx}[1]{\sethlcolor{pink}\hl{[Tong: #1]}}

\title{Cross-device Federated Learning for Mobile Health Diagnostics: \\A First Study on COVID-19 Detection}

\name{Tong Xia, Jing Han, Abhirup Ghosh, Cecilia Mascolo
\thanks{This work was supported by ERC Project 833296 (EAR).}}
\address{Univeristy of Cambridge \\tx229@cam.ac.uk}

\begin{document}
\ninept
\maketitle
\begin{abstract}
Federated learning (FL) aided health diagnostic models can incorporate data from a large number of personal edge devices (e.g., mobile phones) while keeping the data local to the originating devices, largely ensuring privacy. 
However,  such a \emph{cross-device} FL approach for health diagnostics still imposes many challenges due to both local data imbalance (as extreme as local data consists of a single disease class) and global data imbalance (the disease prevalence is generally low in a population). 
Since the federated server has no access to data distribution information, it is not trivial to solve the imbalance issue towards an unbiased model.  
In this paper, we propose \emph{FedLoss}, a novel \emph{cross-device} FL framework for health diagnostics. 
Here the federated server averages the models trained on edge devices according to the predictive loss on the local data, rather than using only the number of samples as weights. As the predictive loss better quantifies the data distribution at a device, \emph{FedLoss}  alleviates the impact of data imbalance.  
Through a real-world dataset on respiratory sound and symptom-based COVID-$19$ detection task, we validate the superiority of \emph{FedLoss}. It achieves competitive COVID-$19$ detection performance compared to a centralised model with an AUC-ROC of $79\%$. It also outperforms the state-of-the-art FL baselines in sensitivity and convergence speed. Our work not only demonstrates the promise of federated COVID-$19$ detection but also paves the way to a plethora of mobile health model development in a privacy-preserving fashion.
\end{abstract}
\section{Introduction}
\vspace{-5pt}
Pervasive mobile devices along with on-device machine learning enable continuous sensing of individual health signals and cost-effective health screening at population scale~\cite{steinhubl2015emerging}. 
However, traditional machine learning methods need the data from all the devices to be aggregated at a central server, raising privacy concerns as the health status and other personally identifiable information can potentially be leaked from the untrusted server or during data sharing~\cite{chauhan2017breathprint}.
Federated learning (FL) avoids aggregating the data and thus promise privacy by iteratively learning models at the participating devices using their local data and then aggregating the local models at a central server~\cite{nguyen2021federated,nguyen2022federated}. This opens a new way for privacy-preserving diagnostic model development.

Most existing diagnostic FL frameworks consider cooperation among hospitals or health institutions with each participant containing clinical data from multiple individuals (also known as \emph{cross-silo} FL setting)~\cite{feki2021federated, qayyum2022collaborative,dou2021federated,yang2021flop}. While such settings have boosted accuracy over participating institutions learning in isolation and improved privacy over centralising the data from all institutes, they still fall short in scaling to more distributed settings where the data of each participant resides on their mobile devices. 
The \emph{cross-silo} FL algorithms do not trivially transfer to \emph{cross-device} FL settings mainly because the latter has many orders of magnitudes more client devices.

\begin{figure}[t]
\centering
\vspace{-15pt}
\includegraphics[width=0.38\textwidth]{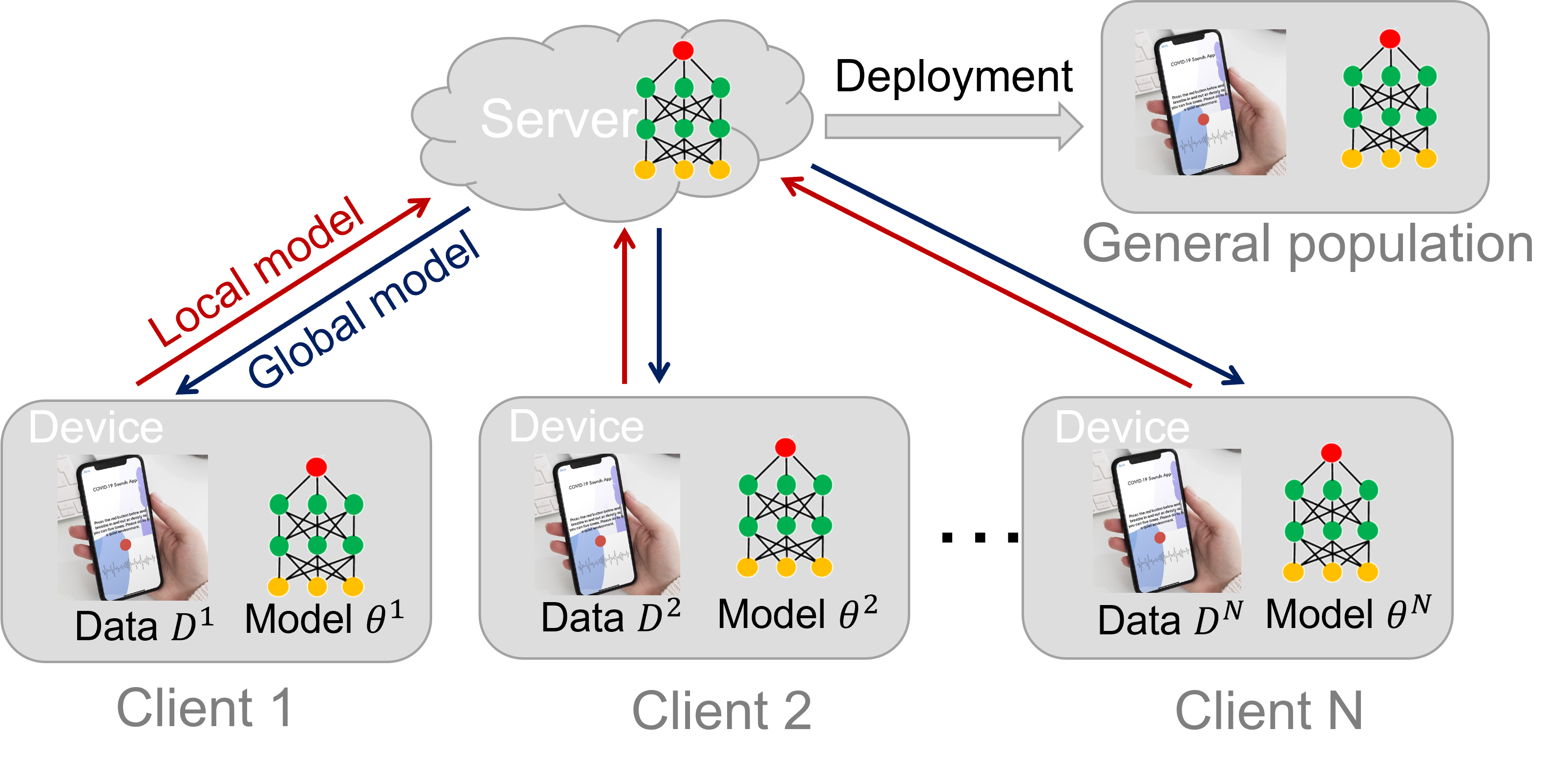}
\vspace{-12pt}
\caption{\emph{Cross-device} FL for mobile health, where models are trained on edge devices from private health sensing data, and the global model is aggregated from the clients' models. 
}\label{fig:fed}
\vspace{-15pt}
\end{figure}

In this paper, we push the envelope of decentralisation by considering \emph{cross-device} FL, where the data resides in users' (clients') edge devices. The learning works in rounds and at every round, each client's edge device trains a model using locally collected health signals and disease labels, while the federated server aggregates the local models into a global one. Finally, the trained model is used for population health screening by any client device using its local sensing data (Fig.~\ref{fig:fed}). 

\emph{Cross-device} FL imposes the following challenges:
i) An individual's health status changes very slowly generally. Therefore, most personal devices will only present a single class, i.e., the current health status of the device owner. It is infeasible to balance the data distribution on the device, and thus learning from such data, the local model is likely to over-fit and be biased.
ii) Due to the generally low disease prevalence, the data is also globally imbalanced, with a large proportion of healthy individuals. Without accessing the label distribution, the global aggregation could introduce an unwanted bias in the classification. 
Yet, failing to detect the disease may come at a heavy price in healthcare applications.

To address the local and global class imbalance, this paper proposes an efficient federated training algorithm, \emph{FedLoss}. The novelty of \emph{FedLoss} lies in its adaptive model aggregation: only a small number of clients are required to participate in each round, and their models are aggregated according to adaptive weights proportional to the predictive loss on their local data. Such an adaptive aggregation strategy alleviates the impact of data imbalance and speeds up global model convergence. 
The performance of \emph{FedLoss} is validated in a COVID-$19$ detection task, where respiratory sounds (cough, breathing, and voice) and symptoms are leveraged to diagnose COVID-$19$.
A dataset is crowd-sourced from around $3,000$ users through a mobile application~\cite{han2021exploring,xia2021covid}. We learn a COVID-$19$ diagnostic classifier where the data stays on the devices, i.e., our experiments consider each user to be a single federated client.

There are two main contributions in this paper. First, we propose a novel federated training algorithm to enable \emph{cross-device} FL for mobile health diagnostics and tackle the challenge resulting from data imbalance. Further, we conduct extensive experiments in a real-world COVID-$19$ detection task. Results demonstrate the superiority of our method over the start-of-the-art baselines.

\vspace{-5pt}
\section{Related Work}
\vspace{-6pt}

Skewed label distribution across edge devices is natural in real-world applications, particularly in the healthcare domain~\cite{rahman2013addressing}. It poses a challenge in FL: due to privacy constraints, class distribution cannot be handled by explicitly identifying the minority class~\cite{shen2021agnostic} and thus it makes the solutions explored in classical centralised settings invalid. Some efforts have concentrated on client clustering~\cite{lin2022fedcluster, sattler2020byzantine}, adapting the global model based on auxiliary data~\cite{wang2021addressing}, and adaptive client training by monitoring the loss from a global perspective~\cite{zhang2021fedpd, shen2021agnostic}. Yet, they either are inefficient when the number of clients is large or require additional centralised data.  A close work to our study~\cite{lin2022fedcluster} (\emph{FedCluster}) considered a \emph{cross-device} setting in FL to diagnose arrhythmia from electrocardiograms. To improve the performance for the rare phenotype, \emph{FedCluster} clusters the clients based on a global shared dataset. Then the local models are first merged within clusters and then cluster models are aggregated into the global model. On the contrary, we aim to solve the imbalance problem without any global data.

\emph{Cross-silo} FL has been explored for health diagnostics including COVID-$19$. For example, Feki~\emph{et al}. proposed FL frameworks allowing multiple medical institutions to screen COVID-$19$ from Chest X-ray images without sharing patient data~\cite{feki2021federated, qayyum2022collaborative,dou2021federated,yang2021flop}. 
Vaid~\emph{et al.} explored electronic medical records to improve mortality prediction across hospitals via FL~\cite{vaid2021federated, dayan2021federated}. In these settings, the number of clients is small and the size of the local data is relatively large. To the best of our knowledge, we are the first to propose a \emph{cross-device} federated learning framework for detecting COVID-$19$ from personal sounds and symptoms. This is more challenging than \emph{cross-silo} FL due to the extreme data heterogeneity from the thousands of clients.
\vspace{-10pt}
\section{Methodology}
\vspace{-6pt}
\subsection{Problem Formulation}
\vspace{-5pt}

Consider a system with $N$ federated clients with each client, $n$ owning a private local dataset $\mathcal{D}^{n}= \{(x^n_1,y^n_1),(x^n_2,y^n_2),...\}$, where $x^n_j$ is a health signal sample and $y^n_j$ denotes the health status, i.e., if the associated disease is identified in the sample, $y^n_j=1$, otherwise $y^n_j=0$.  $y^n_j$ is locally extremely imbalanced with most clients presenting a single class, and it is also globally imbalanced with $y^n_j=0$ (healthy) being the majority class.
As shown in Fig.~\ref{fig:fed}, we aim to train a federated model parameterised by $\theta$ that can predict $y$ for any given $x$  to achieve population health screening.

\vspace{-8pt}
\subsection{Basics of Federated Learning}
\vspace{-5pt}
Federated learning is an iterative process consisting of the following steps at every round:
$(1)$ At every round, $t$, each participating client, $i$ receives a copy of the global model from the previous round, $\theta_{t-1}$ and updates it using its private local data to $\theta_t^i$. $(2)$ Each participating client sends updated  model parameters, $g_t^i = \theta_t^i - \theta_{t-1}$ to the server. $(3)$ The server updates the global model to $\theta_{t}$ by aggregating $g_t^i$s. $(4)$  Steps $(1)$ to $(3)$ are repeated until the global model converges.

The most popular aggregation strategy (step $3$) is Federated Averaging (\emph{FedAvg})~\cite{mcmahan2017communication,li2020federated,gao2022end,feng2022federated}, where the aggregation is an  average of the model updates weighted by $\alpha^i_t$, the fraction of the data samples at client $i$ w.r.t. to the total samples available in the system,
\vspace{-5pt}
\begin{equation}
\theta_{t} = \theta_{t-1} - \eta \sum_i \alpha^i_t g_t^i,
\vspace{-10pt}
\end{equation}
where $\eta$ is the global updating rate.

\begin{algorithm2e}[t]
    \DontPrintSemicolon
    \caption{FedLoss Algorithm}
    \label{al:train}
    \KwData{Global model update rate $\eta$, global training rounds $T$, local update rate $\lambda$, local training epochs $E$, the number of clients each round $M$.} 
    \KwResult{Global model $\theta_T$.}  
    \textbf{Server executes:} \\
    Initialise $\theta_0$ \\
    \For {each round $t$ = 1,2,...,T }  
    { $S_t$ $\leftarrow$ A \ random \ set  \ of \ $M$ \ clients \\
      \For {each client $i \in S_t$ in parallel }  
     {  
       $l_{t}^i, g_{t}^i$ $\leftarrow$  $i$-th \ client \ executes\\
    } 
    $w_{t} = softmax(l_{t}^1,...,l_{t}^M)$ \# Different from FedAvg \\
    $\theta_{t}$ $\leftarrow$ $\theta_{t-1} - \eta \sum_{i=1}^M w_{t}^i g_{t}^i$  
  }
  \hrulefill \\
  \textbf{Client executes:} \\
  Received a global model $\theta_{t-1}$ \\
  Initialise loss $l^i_t$ = 0 \\
  \For {sample $j = 1,2,..., |\mathcal{D}^i|$}  
    { $l^i_t \leftarrow l^i_t + CrossEntropy(\theta_{t-1};\mathcal{D}_j^i$)  \# Returning loss}
    
  Synchronise local model $\theta_{t,0}^i$ =  $\theta_{t-1}$ \\
  \For {local epoch $e = 1,2,...,E$}  
    { $\theta_{t,e}^i \leftarrow  \theta_{t,e-1}^i - \lambda  \nabla_{\theta} CrossEntropy(\theta_{t,e-1}^i;\mathcal{D}^i)$ }
  Calculate the overall update: $g^i_t=\theta_{t,E}^i - \theta_{t-1}$\\
  Return $l_t^i, g_t^i$ \\
\end{algorithm2e} 

\vspace{-5pt}
\subsection{FedLoss}
\vspace{-5pt}
\emph{FedAvg} is vulnerable to class imbalance as $\alpha_t^{i}$ ignores the label imbalance among the clients. 
To overcome this, we propose \emph{FedLoss} (Algorithm~\ref{al:train}) to achieve adaptive aggregation.

At each round of \emph{FedLoss}, $M$ clients are randomly selected to participate in training. Each selected client, $i$, optimises the received model for $E$ epochs using the local data $\mathcal{D}^i$. 
The major difference between \emph{FedLoss} and \emph{FedAvg} is that at each round $t$ in addition to sharing models, client $i$ provides the predictive loss, $l_t^i$ to support a weighted aggregation. $l_t^i$ denotes the total cross-entropy loss incurred by the global model, $\theta_t$ on its local data, $D^i$. Note that $l_t^i$ is computed prior to the local training step and thus it does not suffer from over-fitting at a client with small data. 

Since unhealthy clients are under-represented (globally minority class), intuitively they are more likely to yield relatively higher predictive loss. Thus, \emph{FedLoss} will assign a higher weight to their model updates, rendering the data on such clients to be more predictable by the global model.
Finally, the server normalises the received losses using a \emph{softmax} function to get the client-wise weights for aggregation. The adaptive aggregation in $t$-th round is denoted as,
\begin{equation}
\vspace{-5pt}
\begin{aligned}
    &w_{t} = softmax(l_{t}^1,...,l_{t}^M), \\
    &\theta_{t} =\theta_{t-1}- \eta \sum_{i=1}^M w_{t}^i g_{t}^i,
\end{aligned}
\end{equation} where $w_t^i$ denotes the weight for the participating  client $i$.
The overall process is summarised in Algorithm~\ref{al:train}.

\vspace{-8pt}
\section{Experimental Setup}
\vspace{-6pt}
This section empirically evaluates \emph{FedLoss} for COVID-$19$ detection.
\vspace{-20pt}
\subsection{Data Details}
\begin{figure}[t]
  \centering
  \subfigure[Demographics.]{
  \includegraphics[width=0.225\textwidth]{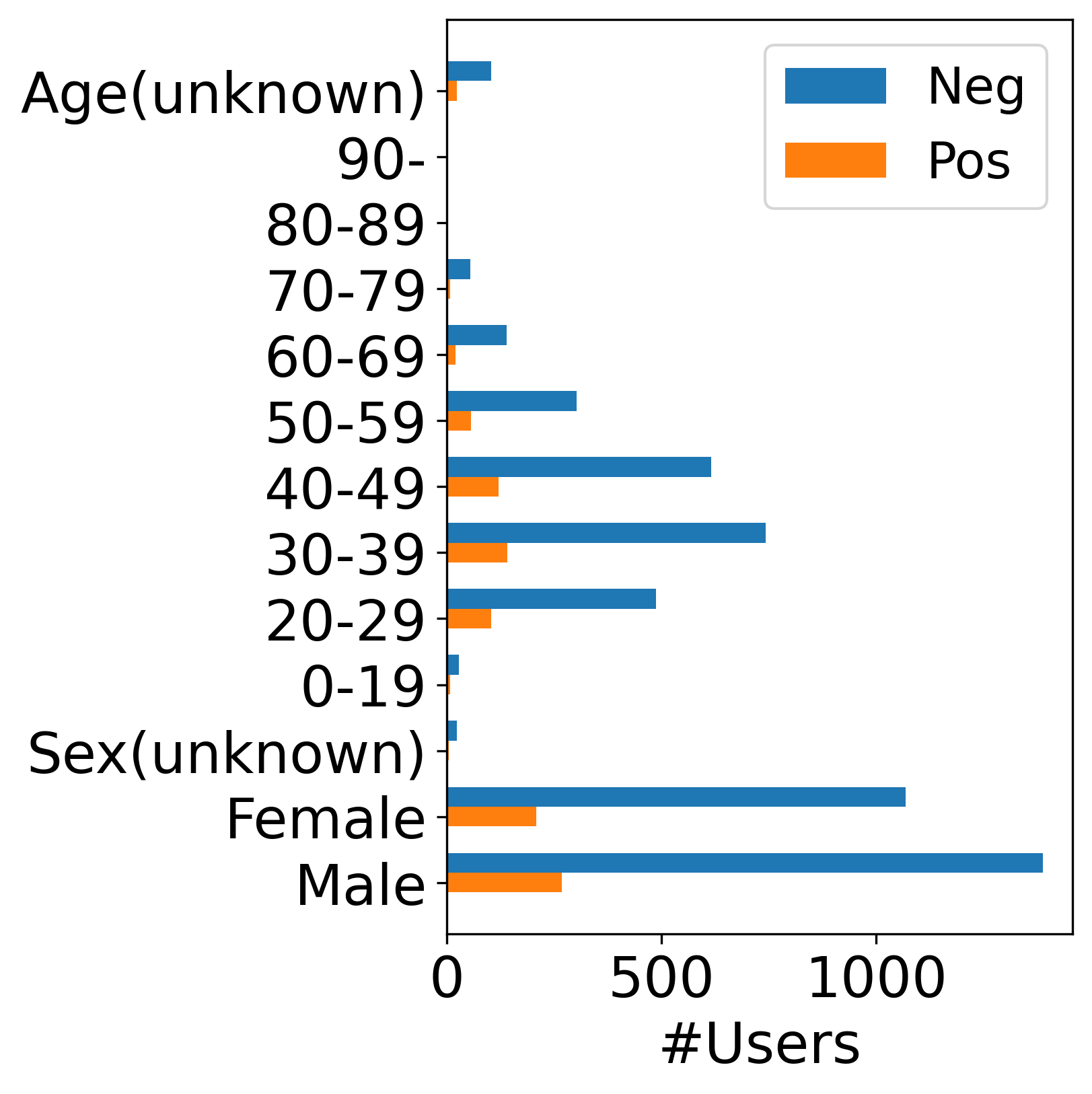}}
  \vspace{-6pt}
  \subfigure[Symptoms distribution.]{
  \includegraphics[width=0.235\textwidth]{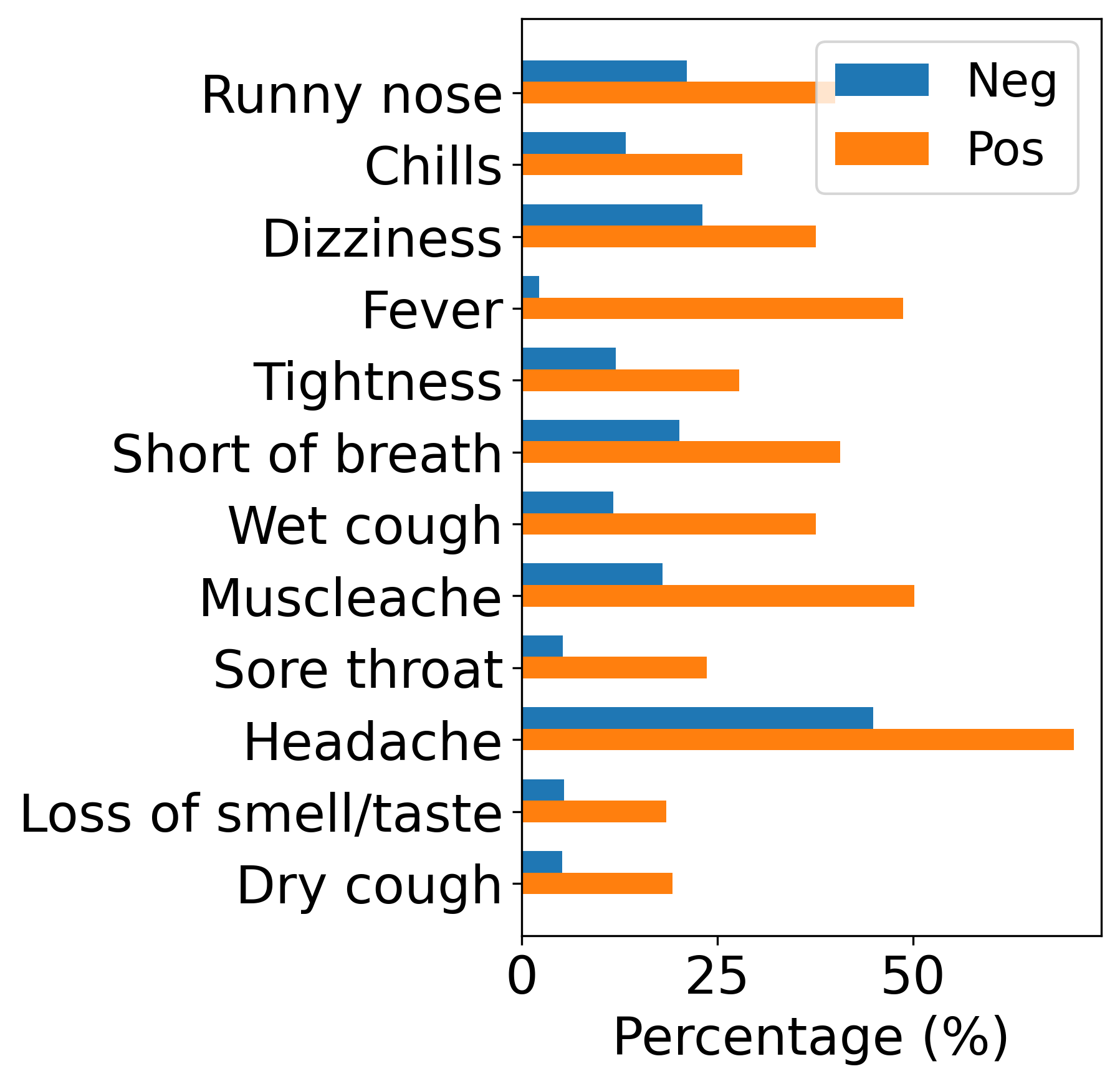}}
  \subfigure[User data amount.]{
  \includegraphics[width=0.215\textwidth]{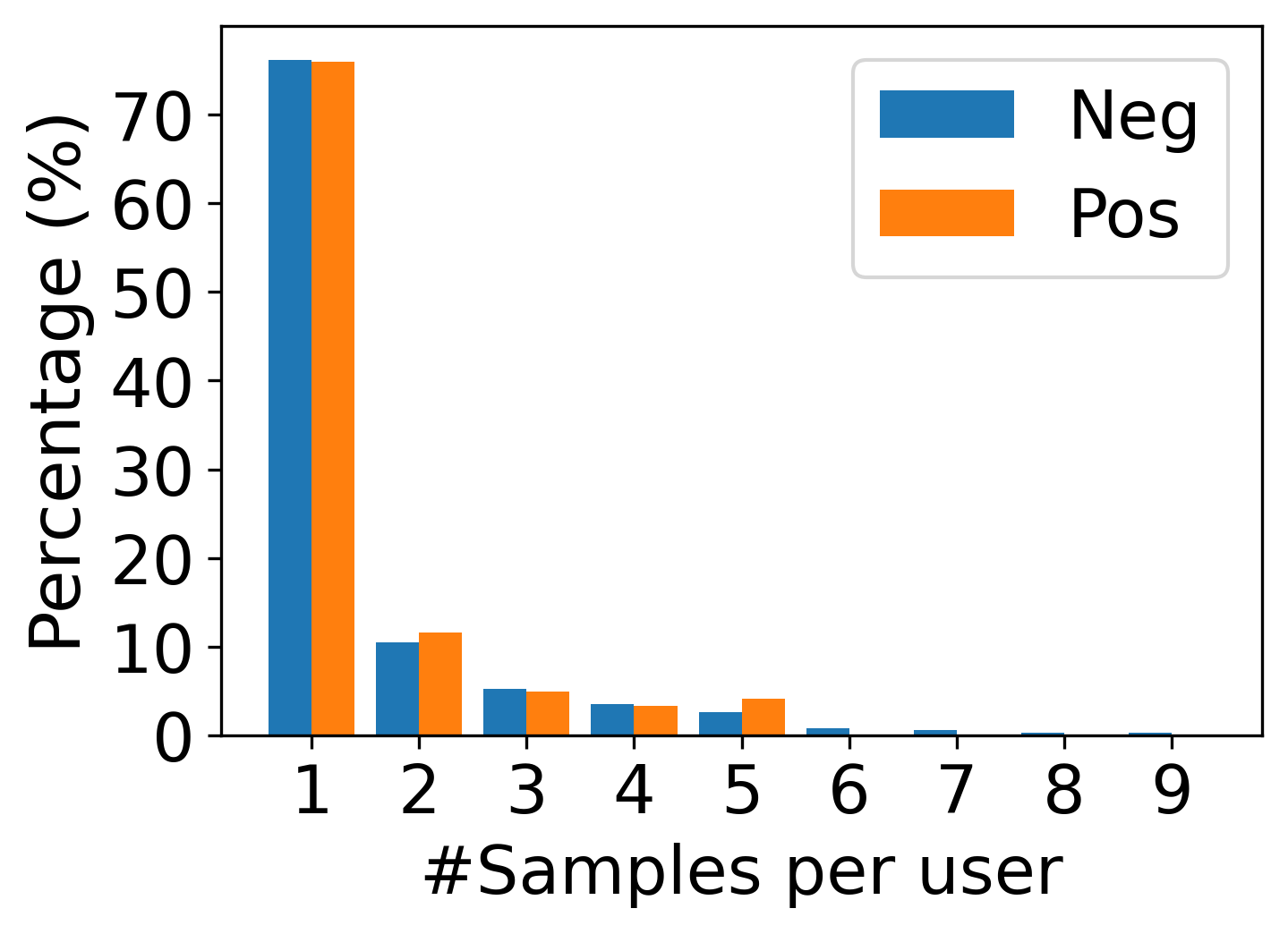}}
  \vspace{-10pt}
  \subfigure[Monthly Data amount.]{
  \includegraphics[width=0.25\textwidth]{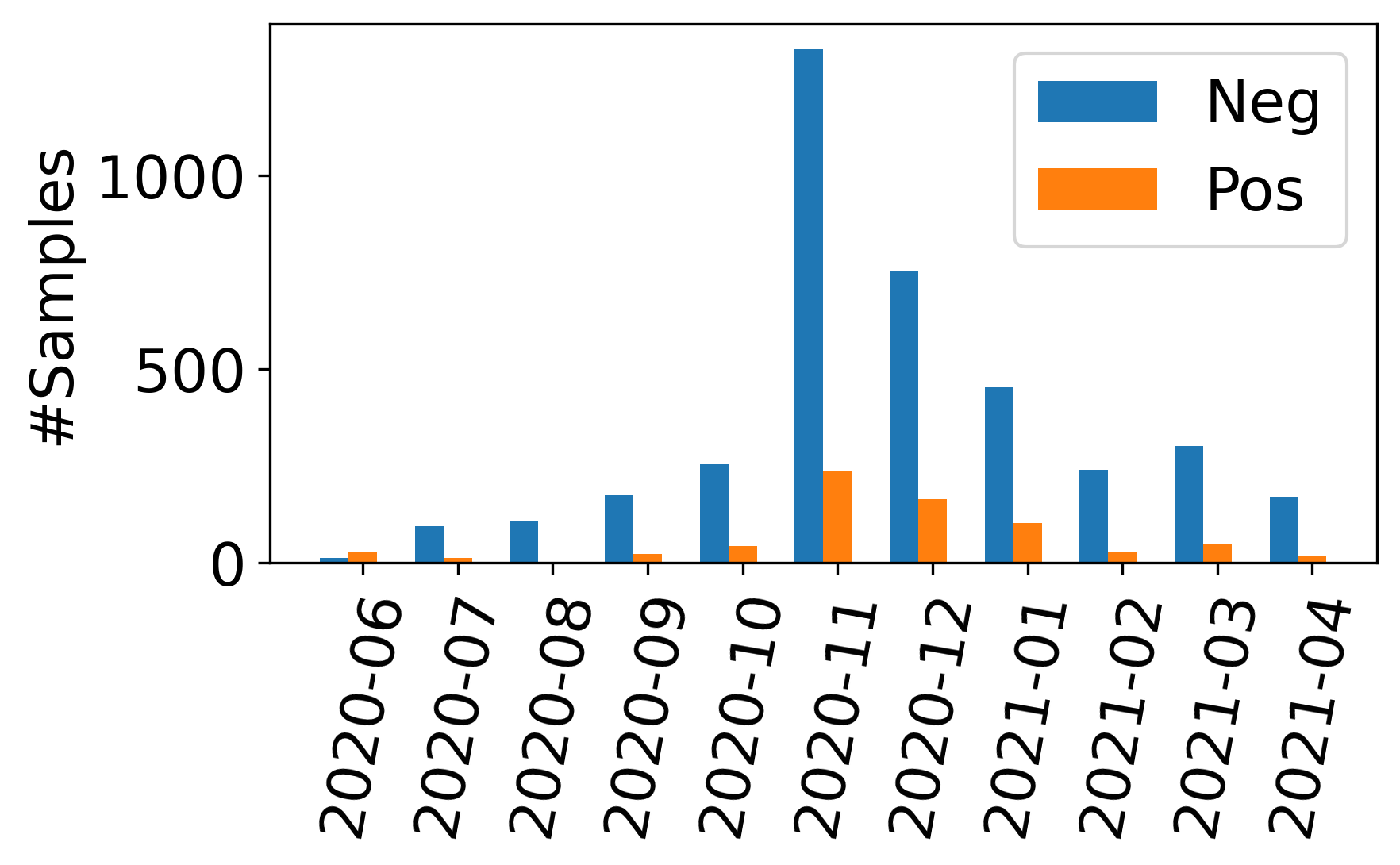}}
  \caption{Statistics of the data from $482$ COVID-19 positive users and  $2,478$  negative users.}
  \label{fig:data}
   \vspace{-10pt}
\end{figure}

\vspace{-6pt}
We use the data collected by a crow-sourced mobile application, \emph{COVID-19 Sounds}\footnote{\url{www.covid-19-sounds.org}}.
At registration, the app assigns each user a unique anonymous ID. Users record their symptoms (cough, fever, etc.), three respiratory sound recordings (breathing, coughing, and speech), and the COVID-$19$ testing status on the corresponding day~\cite{han2021exploring,xia2021covid}. 
After data cleaning (i.e., excluding non-English speakers, samples without COVID-$19$ test results and poor audio quality samples),  there are $482$ users with positive status and $2,478$ users with negative status with a total of $4,612$ samples. An overview of the statistics of the data is in Fig.~\ref{fig:data}: (a) The data represents a typical demographic distribution in a population. (b) There are more negative than positive users, with many asymptomatic positive users while a great proportion of the negative users report respiratory disease-related symptoms. (c) User data is sparse with over $70\%$ of users only recording one sample. (d) The data accumulation procession spanned one year.
 

\vspace{-10pt}
 \subsection{Backbone COVID-19 Detection Model}
\begin{figure}[t]
\centering
\includegraphics[width=0.4\textwidth]{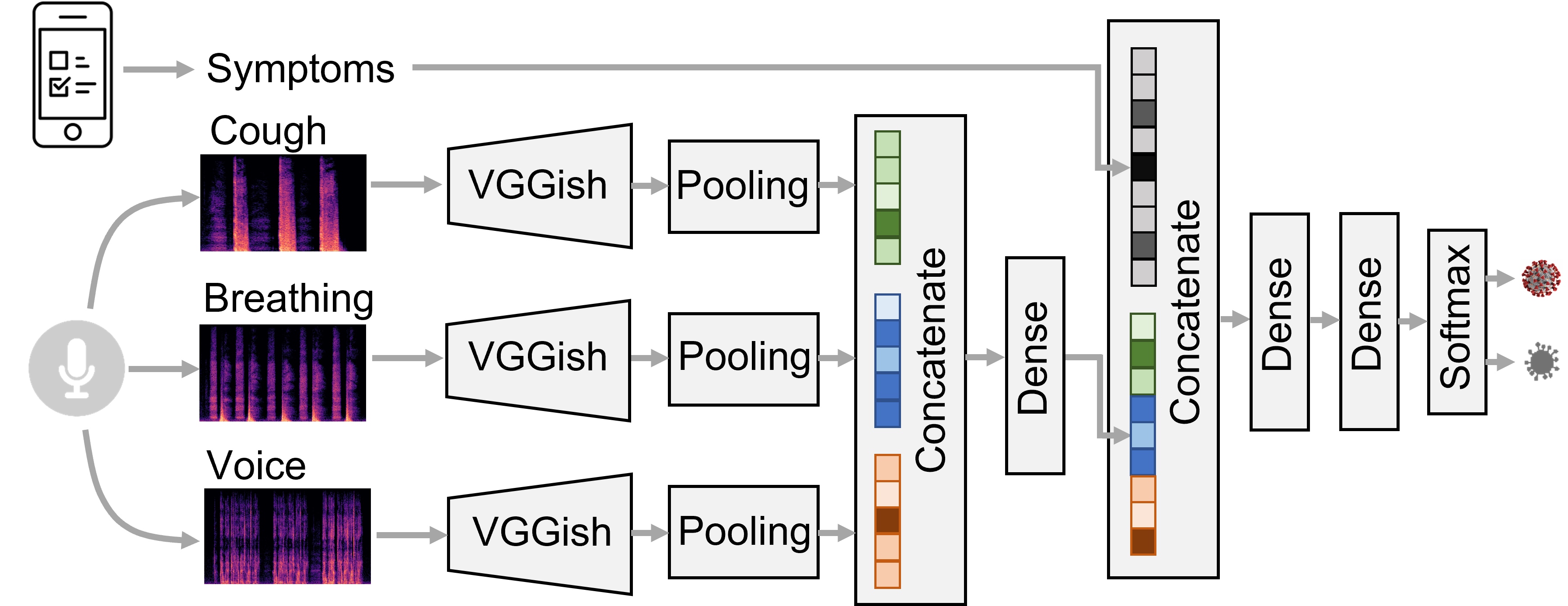}
\caption{\label{fig:model} A multi-modal model for COVID-19 detection.}
 \vspace{-10pt}
\end{figure}

 \vspace{-6pt}
Following the previous works~\cite{xia2021covid,han2022sounds}, a VGGish framework is employed to extract acoustic features from the spectrogram of audio samples. Additionally, Han \emph{et al.} reported that fusing the symptoms and acoustic features in an early stage of the deep model can achieve better COVID-$19$ detection performance than using a single modality. Inspired by this, we use a multi-modal deep learning model to predict COVID-$19$ status from audio and symptoms jointly, as illustrated in Fig.~\ref{fig:model}. Symptoms are represented by a multi-hot vector, which is concatenated with the dense feature from VGGish network outputs. The concatenated feature vector is then fed to a multi-layer fully connected network for classification. The final layer outputs a \emph{Softmax} based binary class probabilities.

\vspace{-5pt}
\subsection{Settings}
\vspace{-2pt}
This paper considers each app user as a federated client to examine \emph{FedLoss}. Out of $2,960$ users in the dataset we randomly held out $20\%$ clients for testing and use the rest $80\%$ of the clients for federated training. We experiment with two training settings:
\begin{itemize}
    \item \textbf{Randomly}: The recorded data is assumed to be kept on the client device during the whole training period. We run $T=2000$ federated rounds and $M=30$ clients are randomly selected at each round.
    \item \textbf{Chronologically}: The recorded data is assumed to be cleared monthly by the user, which is practical.  Regarding this, we design a multi-period training strategy: every month, only the clients with data recorded in this period can be selected and we run $100$ rounds with each round sampling $M=30$ clients for training ($100$ rounds can guarantee the convergence of the model on the incremental data).
\end{itemize}
 All the experiments are implemented by Pytorch on a GPU with 64G memory. To avoid over-fitting on the client, a pre-trained VGGish is utilised~\cite{xia2021covid}, and the local training epoch is set to $E=1$.  A local learning rate of $0.008$ for VGGish and $0.015$ for the rest parameters are used for the SGD optimiser. The global update rate $\eta=1$.

\vspace{-8pt}
\subsection{Baselines and Metrics}
\vspace{-5pt}
In addition to \emph{FedAvg}, we also compare with \emph{FedProx}~\cite{li2020federated}. \emph{FedProx} handles non-identically distributed data across federated clients by regularising the local training loss at the clients so that the local models incur limited  divergence from the global model. 

For evaluation, we first use AUC-ROC (short for AUC) to show the overall rationality of the estimated diagnostic probability. 
Following the rule that for a sample if the predictive probability of  being positive  is larger than being negative, i.e., $p_{pos} > p_{neg}$, it will be diagnosed as positive, we also present sensitivity (SE) - the ratio between the correctly identified COVID-$19$ positive samples and overall positive samples,  and specificity (SP)  - the correct ratio for the healthy class.
Additionally, we report sensitivity with a specificity of $80\%$ (SE@80\%SP) by tuning the decision threshold, i.e., a sample will only be diagnosed as positive when $p_{pos} > p_{neg} + \tau$, where $\tau$ is searched to guarantee a SP of 80\%.
A 95\% Confidence Interval (CI) for all metrics is reported by using bootstrap~\cite{diciccio1996bootstrap}.

\vspace{-5pt}
\section{Results}
\vspace{-8pt}
\subsection{Results and Discussion under Randomly Training Setting}
\vspace{-4pt}
\textit{\textbf{COVID-19 Detection Performance}}. The overall performance comparison is summarised in Table~\ref{tab:results1}. All federated learning based approaches achieve competitive AUC-ROC against centralised training. However, the federated baselines are unable to effectively detect COVID-$19$ positive users with sensitivity lower than $20\%$, although their specificity is very high. In contrast, our \emph{FedLoss} yields a sensitivity of 50\% while maintaining the specificity around 90\%. In other words, \emph{FedLoss} achieves the best trade-off for detecting positive and negative users, as proved by the highest average value of sensitivity and specificity (70\%).
 When fixing the specificity of 80\% uniformly, our \emph{FedLoss} achieves sensitivity up to 62\%, which is as good as the centralised model. 
 All those validate the superiority of our weighted aggregation strategy in handling the data imbalance.

\begin{table}[t]
\caption{Performance comparison under \emph{randomly} training setting. $95\%$~CIs are reported in brackets.}
\centering
\resizebox{0.43\textwidth}{!}{%
\begin{tabular}{cccccc}
\toprule
    & \textbf{AUC}     & \textbf{SE}   & \textbf{SP} & \textbf{SE@80\%SP}    \\ \midrule
\multirow{2}{*}{\textbf{Centralised}} & $0.79$&$0.46$&$0.93$&$0.62$\\
& $(0.74-0.84)$&$(0.36-0.56)$&$(0.91-0.94)$&$(0.54-0.69)$\\\hline
\multirow{2}{*}{\textbf{FedAvg}} & $0.80$ & $0.11$ & $1.00$ &$0.59$\\
& $(0.75-0.85)$ & $(0.06-0.17)$ & $(1.00-1.00)$ &$(0.45-0.73)$\\\hline
\multirow{2}{*}{\textbf{FedProx}} & $0.75$&$0.19$ & $0.99$ & $0.48$\\ 
& $(0.69-0.80)$ &$(0.12-0.27)$ & $(0.99-1.00)$ & $(0.31-0.63)$\\ \hline
 \textbf{FedLoss}&$0.79$&$0.50$&$0.90$&$0.62$\\ %
 (Proposed)&$(0.73-0.83)$&$(0.40-0.59)$&$(0.88-0.92)$&$(0.50-0.70)$\\ 
\bottomrule
\end{tabular}}
\label{tab:results1}
\vspace{-5pt}
\end{table}


\begin{figure}[t]
\centering
\includegraphics[width=0.4\textwidth]{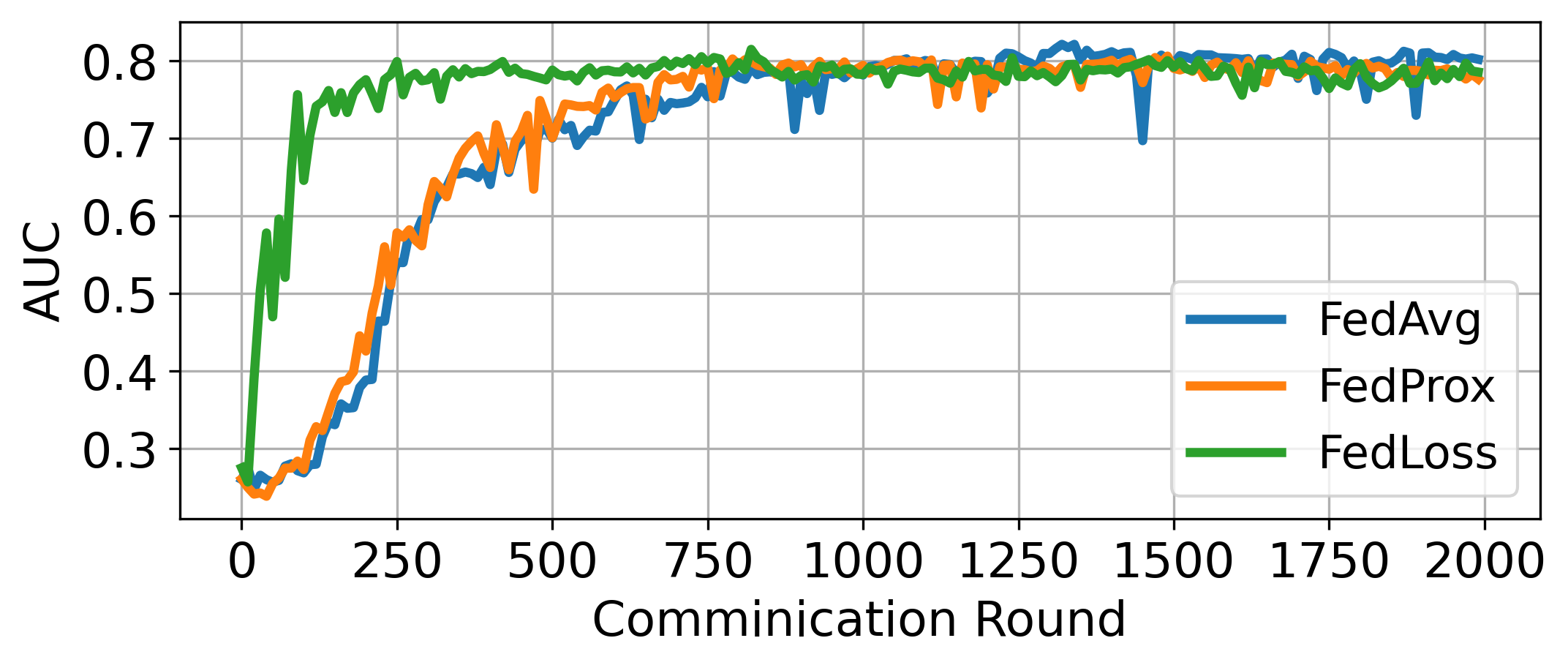}
\vspace{-8pt}
\caption{Convergence analysis. AUC-ROC of testing set for every 10 round during training is displayed.}\label{fig:con} 
\vspace{-8pt}
\end{figure}

\begin{figure}[t]
\centering
\includegraphics[width=0.4\textwidth]{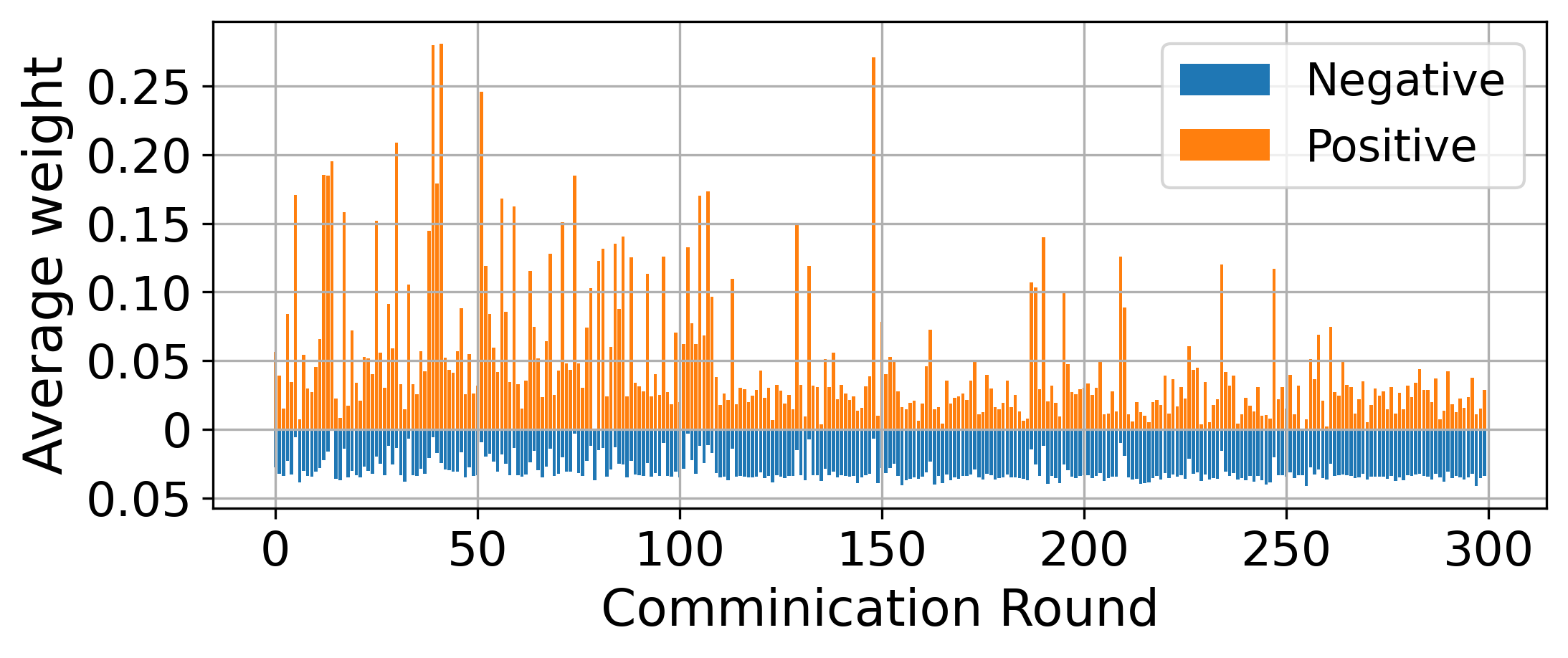}
\vspace{-8pt}
\caption{ Average weight for COVID-19 positive and negative clients per communication round. Note that the negative clients do not have negative weights, but the weights are just shown in negative direction for visualisation convenience. } \label{fig:weight}
\vspace{-10pt}
\end{figure}

\vspace{5pt}\noindent \textit{\textbf{Convergence Comparison}}. System efficiency is another important metric for \emph{cross-device} FL. To compare the convergence speed of \emph{FedAvg},  \emph{FedProx} and \emph{FedLoss}, we show the testing AUC-ROC during the training process in Fig.~\ref{fig:con}. It can be observed that the AUC-ROC of our \emph{FedLoss} gets converged significantly faster than the baselines: \emph{FedLoss} needs about $250$ rounds while \emph{FedAvg} and \emph{FedProx} requires about $1000$ rounds. Therefore, \emph{FedLoss} is $4\times$ more efficient than baselines. Note that fewer communication rounds to convergence saves both computation and communication costs at the resource constraint edge clients.

\vspace{5pt}\noindent \textit{\textbf{Analysis of Weights}}. We conduct additional analysis on the adaptive weight during the training process. Since our \emph{FedLoss} shows a superior sensitivity against the baselines, we particularly look at how the weights changed for COVID-19 positive and negative clients, for a comparison. Fig.~\ref{fig:weight} displays the average weight for positive and negative clients in each round. It is observed that in the beginning 100 rounds, the weight of positive clients is 4$\sim$6 times of negative clients. This suggests the system can detect the potentially minority class as those clients are more difficult to predict. In the later rounds, the weights for positive and negative clients gradually become more balanced, since the global model has already learned the COVID-19 features, to a great extent. 


\vspace{-6pt}
\subsection{Discussion under Chronologically Training Setting}
\begin{figure}[t]
\centering
\includegraphics[width=0.39\textwidth]{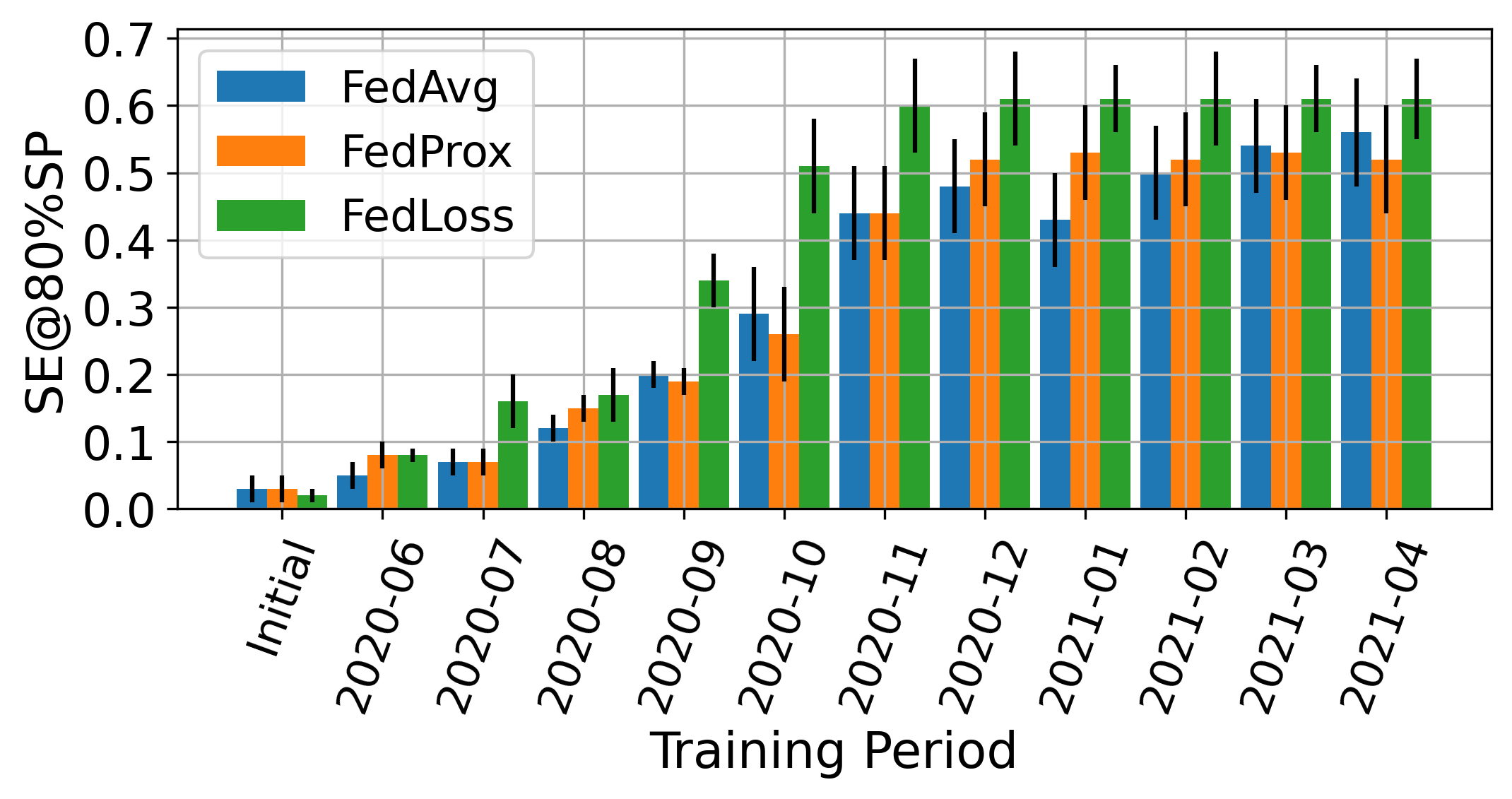}
\vspace{-9pt}
\caption{Performance of the global model trained \emph{chronologically}. Sensitivity with a specificity of 80\% by the last round model in each month is displayed.} \label{fig:chro}
\vspace{-10pt}
\end{figure}

\vspace{-5pt}
The second setting aims to evaluate the performance of long-term FL with limited client participation in batches.  
As illustrated by the SE@80\%SP in different periods in Fig.~\ref{fig:chro}, all methods are inaccurate and unusable at the early stage with SE@80\%SP lower than $50\%$. The poor performance is mainly attributed to the limited number of clients (i.e., the limited data), which leads to poor generalisation.
Gradually, with more training rounds, from November 2020 our \emph{FedLoss} starts to show a convergence trend with the SE@80\%SP reaching $60\%$. Finally, our model achieves an AUC-ROC of $79\%$, a sensitivity of $45\%$ and specificity of $90\%$, as summarised in Table~\ref{tab:results2}. 
On the contrary,  SE@80\%SP of \emph{FedAvg} and \emph{FedProx} has slower convergence rate, converging two months later than \emph{FedLoss}. We also note that in November 2020, all three approaches present a remarkable performance gain, which is mainly because the quantity of data reaches a peak in that month (refer to Fig.~\ref{fig:data}(d)).
Overall, our final SE@80\%SP ($62\%$) significantly surpasses that of \emph{FedAvg} ($56\%$) and \emph{FedProx} ($53\%$), and our SE ($45\%$) is quite competitive compared with centralised model ($46\%$).
The above comparison further verifies that our proposed \emph{FedProx} can achieve a more generalised global model with fewer clients involved.

\begin{table}[!t]
\caption{Overall performance of the final model under \emph{chronologically} training setting. $95\%$~CIs are reported in brackets.}
\centering
\resizebox{0.42\textwidth}{!}{%
\begin{tabular}{cccccc}
\toprule
    & \textbf{AUC}     & \textbf{SE}   & \textbf{SP} & \textbf{SE@80\%SP}    \\ \midrule
\multirow{2}{*}{\textbf{FedAvg}} & $0.79$&$0.20$&$0.99$&$0.56$\\
&$(0.73-0.82)$&$(0.15-0.23)$&$(0.98-1.00)$&$(0.49-0.63)$\\\hline
\multirow{2}{*}{\textbf{FedProx}} &$0.78$&$0.15$&$0.99$&$0.53$\\ 
&$(0.75-0.81)$&$(0.10-0.20)$&$(0.98-1.00)$&$(0.44-0.60)$\\ \hline
\textbf{FedLoss}  &$0.79$&$0.45$&$0.90$&$0.61$\\ %
(Proposed)&$(0.74-0.84)$&$(0.39-0.53)$&$(0.89-1.0)$&$(0.55-0.64)$\\
\bottomrule
\end{tabular}}
\label{tab:results2}
\vspace{-8pt}
\end{table}
\vspace{-5pt}
\section{Conclusion}
\vspace{-5pt}
In this paper, we studied the feasibility of \emph{cross-device} federated mobile health using a COVID-$19$ detection task as an example. To handle the natural challenge of data imbalance, a novel federated aggregation algorithm \emph{FedLoss} has been proposed. Experimental results demonstrate the superiority of our approach in both effectiveness and efficiency. \emph{FedLoss} aggregation scheme is general and can be extended to other mobile health applications, e.g., heart sound-based arrhythmia prediction, and smartwatch-enabled sleep quality monitoring. This paper also facilitates the change from traditional crowdsourcing of data to crowdsourcing of models on a large scale for privacy-preserving mobile health applications. While this study is a beginning of an exciting direction of \emph{cross-device} federated mobile health, many challenges lie ahead, for example, the sparsity of labelled data at the devices, addressing which will be future work.


\clearpage
\balance

 
\bibliographystyle{IEEEbib}
\bibliography{refs}

\begin{thebibliography}{10}

\bibitem{steinhubl2015emerging}
Steven~R Steinhubl, Evan~D Muse, and Eric~J Topol,
\newblock ``The emerging field of mobile health,''
\newblock {\em Science Translational Medicine}, vol. 7, no. 1, pp. 283--295,
  2015.

\bibitem{chauhan2017breathprint}
Jagmohan Chauhan, Yining Hu, Suranga Seneviratne, Archan Misra, Aruna
  Seneviratne, and Youngki Lee,
\newblock ``Breathprint: Breathing acoustics-based user authentication,''
\newblock in {\em Proc. International Conference on Mobile Systems,
  Applications, and Services}, 2017, pp. 278--291.

\bibitem{nguyen2021federated}
Dinh~C Nguyen, Ming Ding, Pubudu~N Pathirana, Aruna Seneviratne, Jun Li, and
  H~Vincent Poor,
\newblock ``Federated learning for internet of things: A comprehensive
  survey,''
\newblock {\em IEEE Communications Surveys \& Tutorials}, vol. 23, no. 3, pp.
  1622--1658, 2021.

\bibitem{nguyen2022federated}
Dinh~C Nguyen, Quoc-Viet Pham, Pubudu~N Pathirana, Ming Ding, Aruna
  Seneviratne, Zihuai Lin, Octavia Dobre, and Won-Joo Hwang,
\newblock ``Federated learning for smart healthcare: A survey,''
\newblock {\em ACM Computing Surveys}, vol. 55, no. 3, 2022,
\newblock 37 pages.

\bibitem{feki2021federated}
Ines Feki, Sourour Ammar, Yousri Kessentini, and Khan Muhammad,
\newblock ``Federated learning for {COVID}-19 screening from chest {X-Ray}
  images,''
\newblock {\em Applied Soft Computing}, vol. 106, pp. 107330, 2021.

\bibitem{qayyum2022collaborative}
Adnan Qayyum, Kashif Ahmad, Muhammad~Ahtazaz Ahsan, Ala Al-Fuqaha, and Junaid
  Qadir,
\newblock ``Collaborative federated learning for healthcare: Multi-modal
  {COVID}-19 diagnosis at the edge,''
\newblock {\em IEEE Open Journal of the Computer Society}, vol. 1, pp.
  172--184, 2022.

\bibitem{dou2021federated}
Qi~Dou, Tiffany~Y So, Meirui Jiang, Quande Liu, Varut Vardhanabhuti, Georgios
  Kaissis, Zeju Li, Weixin Si, Heather~HC Lee, Kevin Yu, et~al.,
\newblock ``Federated deep learning for detecting {COVID}-19 lung abnormalities
  in {CT}: a privacy-preserving multinational validation study,''
\newblock {\em NPJ Digital Medicine}, vol. 4, no. 1, 2021,
\newblock 11 pages.

\bibitem{yang2021flop}
Qian Yang, Jianyi Zhang, Weituo Hao, Gregory~P Spell, and Lawrence Carin,
\newblock ``Flop: Federated learning on medical datasets using partial
  networks,''
\newblock in {\em Proc. ACM Conference on Knowledge Discovery \& Data Mining
  (SIGKDD)}, 2021, pp. 3845--3853.

\bibitem{han2021exploring}
Jing Han, Chlo{\"e} Brown, Jagmohan Chauhan, Andreas Grammenos, Apinan
  Hasthanasombat, Dimitris Spathis, Tong Xia, Pietro Cicuta, and Cecilia
  Mascolo,
\newblock ``Exploring automatic {COVID}-19 diagnosis via voice and symptoms
  from crowdsourced data,''
\newblock in {\em Proc. IEEE International Conference on Acoustics, Speech and
  Signal Processing (ICASSP)}, 2021, pp. 8328--8332.

\bibitem{xia2021covid}
Tong Xia, Dimitris Spathis, Jagmohan Chauhan, Andreas Grammenos, Jing Han,
  Apinan Hasthanasombat, Erika Bondareva, Ting Dang, Andres Floto, Pietro
  Cicuta, et~al.,
\newblock ``{COVID}-19 sounds: A large-scale audio dataset for digital
  respiratory screening,''
\newblock in {\em Proc. Conference on Neural Information Processing Systems
  Datasets and Benchmarks Track}, 2021,
\newblock 10 pages.

\bibitem{rahman2013addressing}
M~Mostafizur Rahman and Darryl~N Davis,
\newblock ``Addressing the class imbalance problem in medical datasets,''
\newblock {\em International Journal of Machine Learning and Computing}, vol.
  3, no. 2, pp. 224, 2013.

\bibitem{shen2021agnostic}
Zebang Shen, Juan Cervino, Hamed Hassani, and Alejandro Ribeiro,
\newblock ``An agnostic approach to federated learning with class imbalance,''
\newblock in {\em Proc. International Conference on Learning Representations
  (ICLR)}, 2021.

\bibitem{lin2022fedcluster}
Daoqin Lin, Yuchun Guo, Huan Sun, and Yishuai Chen,
\newblock ``Fedcluster: A federated learning framework for cross-device private
  ecg classification,''
\newblock in {\em Proc. IEEE Conference on Computer Communications Workshops
  (INFOCOM WKSHPS)}, 2022,
\newblock 6 pages.

\bibitem{sattler2020byzantine}
Felix Sattler, Klaus-Robert M{\"u}ller, Thomas Wiegand, and Wojciech Samek,
\newblock ``On the byzantine robustness of clustered federated learning,''
\newblock in {\em Proc. IEEE International Conference on Acoustics, Speech and
  Signal Processing (ICASSP)}, 2020, pp. 8861--8865.

\bibitem{wang2021addressing}
Lixu Wang, Shichao Xu, Xiao Wang, and Qi~Zhu,
\newblock ``Addressing class imbalance in federated learning,''
\newblock in {\em Proc. AAAI}, 2021, vol.~35, pp. 10165--10173.

\bibitem{zhang2021fedpd}
Xinwei Zhang, Mingyi Hong, Sairaj Dhople, Wotao Yin, and Yang Liu,
\newblock ``Fedpd: A federated learning framework with adaptivity to non-iid
  data,''
\newblock {\em IEEE Transactions on Signal Processing}, vol. 69, pp.
  6055--6070, 2021.

\bibitem{vaid2021federated}
Akhil Vaid, Suraj~K Jaladanki, Jie Xu, Shelly Teng, Arvind Kumar, Samuel Lee,
  Sulaiman Somani, Ishan Paranjpe, Jessica~K De~Freitas, Tingyi Wanyan, et~al.,
\newblock ``Federated learning of electronic health records to improve
  mortality prediction in hospitalized patients with {COVID}-19: machine
  learning approach,''
\newblock {\em JMIR Medical Informatics}, vol. 9, no. 1, pp. e24207, 2021.

\bibitem{dayan2021federated}
Ittai Dayan, Holger~R Roth, Aoxiao Zhong, Ahmed Harouni, Amilcare Gentili,
  Anas~Z Abidin, Andrew Liu, Anthony~Beardsworth Costa, Bradford~J Wood,
  Chien-Sung Tsai, et~al.,
\newblock ``Federated learning for predicting clinical outcomes in patients
  with {COVID}-19,''
\newblock {\em Nature Medicine}, vol. 27, no. 10, pp. 1735--1743, 2021.

\bibitem{mcmahan2017communication}
Brendan McMahan, Eider Moore, Daniel Ramage, Seth Hampson, and Blaise~Aguera
  y~Arcas,
\newblock ``Communication-efficient learning of deep networks from
  decentralized data,''
\newblock in {\em Proc. Artificial Intelligence and Statistics}, 2017, pp.
  1273--1282.

\bibitem{li2020federated}
Tian Li, Anit~Kumar Sahu, Ameet Talwalkar, and Virginia Smith,
\newblock ``Federated learning: Challenges, methods, and future directions,''
\newblock {\em IEEE Signal Processing Magazine}, vol. 37, no. 3, pp. 50--60,
  2020.

\bibitem{gao2022end}
Yan Gao, Titouan Parcollet, Salah Zaiem, Javier Fernandez-Marques, Pedro~PB
  de~Gusmao, Daniel~J Beutel, and Nicholas~D Lane,
\newblock ``End-to-end speech recognition from federated acoustic models,''
\newblock in {\em Proc. IEEE International Conference on Acoustics, Speech and
  Signal Processing (ICASSP)}, 2022, pp. 7227--7231.

\bibitem{feng2022federated}
Meng Feng, Chieh-Chi Kao, Qingming Tang, Ming Sun, Viktor Rozgic, Spyros
  Matsoukas, and Chao Wang,
\newblock ``Federated self-supervised learning for acoustic event
  classification,''
\newblock in {\em Proc. IEEE International Conference on Acoustics, Speech and
  Signal Processing (ICASSP)}, 2022, pp. 481--485.

\bibitem{han2022sounds}
Jing Han, Tong Xia, Dimitris Spathis, Erika Bondareva, Chlo{\"e} Brown,
  Jagmohan Chauhan, Ting Dang, Andreas Grammenos, Apinan Hasthanasombat, Andres
  Floto, et~al.,
\newblock ``Sounds of {COVID}-19: Exploring realistic performance of
  audio-based digital testing,''
\newblock {\em NPJ Digital Medicine}, vol. 5, no. 1, 2022,
\newblock 9 pages.

\bibitem{diciccio1996bootstrap}
Thomas~J DiCiccio and Bradley Efron,
\newblock ``Bootstrap confidence intervals,''
\newblock {\em Statistical Science}, vol. 11, no. 3, pp. 189--228, 1996.

\end{thebibliography}
\end{document}